\documentclass[conference, lettersize]{IEEEtran}
\IEEEoverridecommandlockouts
\usepackage{cite}
\usepackage{amsmath,amssymb,amsfonts}
\usepackage{hyperref}
\usepackage{graphicx}
\usepackage{placeins}
\usepackage{booktabs}
\usepackage{textcomp}
\usepackage{booktabs}
\usepackage{xcolor}
\usepackage{float}
\usepackage{bm}
\usepackage{derivative}
\usepackage{bbm}
\usepackage{subcaption}

\hypersetup{
    colorlinks,
    linkcolor={red!50!black},
    citecolor={blue!50!black},
    urlcolor={blue!80!black}
}

\def\BibTeX{{\rm B\kern-.05em{\sc i\kern-.025em b}\kern-.08em
    T\kern-.1667em\lower.7ex\hbox{E}\kern-.125emX}}

\begin{document}

\title{Fine color guidance in diffusion models and its application to image compression at extremely low bitrates}

\author{\IEEEauthorblockN{Tom Bordin}\IEEEauthorblockA{INRIA\\Rennes, France}\\
\and
\IEEEauthorblockN{Thomas Maugey}\IEEEauthorblockA{INRIA\\Rennes France}
}

\maketitle

\begin{abstract}

This study addresses the challenge of, without training or fine-tuning, controlling the global color aspect of images generated with a diffusion model. We rewrite the guidance equations to ensure that the outputs are closer to a known color map, and this without hindering the quality of the generation. Our method leads to new guidance equations. We show in the color guidance context that, the scaling of the guidance should not decrease but remains high throughout the diffusion process. In a second contribution, our guidance is applied in a compression framework, we combine both semantic and general color information on the image to decode the images at low cost. We show that our method is effective at improving fidelity and realism of compressed images at extremely low bit rates, when compared to other classical or more semantic oriented approaches. 

\end{abstract}

\section{Introduction}

Image generation has made significant progress, mainly due to the emergence of diffusion models\cite{song2019generative}. Those generative methods rely heavily on computations and data in order to produce the highest image quality. Therefore, when using the most recent networks for other applications, it can be interesting to avoid retraining or fine-tuning it but rather adapt the models. The current popular and widely available conditioning, done through text or image, is often not enough for any applications that require finer control. Getting control on the outputs without retraining thus becomes a challenge. One of the difficulty of controlling image generation lies in the fact that most recent models are trained to work in the latent space of a fixed VAE, as proposed in \cite{rombach_high-resolution_2022}, those models are called Latent Diffusion Models(LDMs). By reducing the dimension considered, the model can be shrunk, and the generation process can be sped up, but this comes at the expense of interpretability and control. The solutions for controlling outputs in pixel space are not always relevant for LDMs, while unfortunately most current state-of-the-art models are trained in this configuration.

\begin{figure}[htbp]
    \centering
    \includegraphics[width=0.49\textwidth]{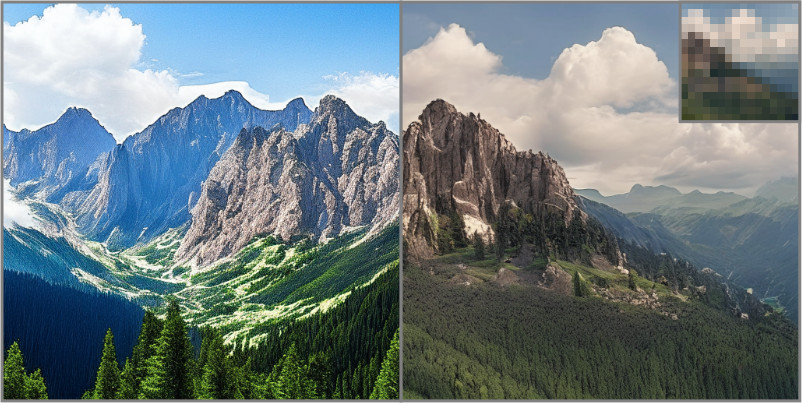}
    \caption{Two images generated with a Diffusion Model, conditioned by the same semantics (\textit{rocky mountain and cloudy sky}). Contrary to the left image, the right image is obtained by guiding the diffusion model towards a general color aspect (top right).}
    \label{fig:intro}
\end{figure}

In this paper, we focus on a specific type of control on LDMs, we aim at generating an image that is coherent with a given general color aspect which is given by a color map, we illustrate the concept in Fig. \ref{fig:intro}, adding control to an already semantic conditional generation. Control through color maps can be a useful tool for image generation, especially in the context of compression, where, a really coarse color map can be used as a base to generate details. There already exist some options available to control color for diffusion in the pixel space, such as generating only the high frequencies using a decomposition of the image proposed by the authors of \cite{you2023indigo}, or using a specific scheduler for the diffusion model \cite{kawar2022denoising}. However, to the best of our knowledge, a good control of the color in LDM without training, has not been presented. Some works propose training methods to add control to diffusion models, by learning the directions in the latent space associated with semantic meaning\cite{kwon2022diffusion} or by adding a condition as input\cite{zhang2023adding}. However, we do not want to rely on training or fine-tuning a new model on the specific task of color map conditioning, as this would not take advantage of the most recent models that could be trained in the future. We want to focus on a training-free control that is adaptable to any diffusion model, and in particular to LDMs. In SdEdit\cite{meng2021sdedit}, the authors propose a method that provides some control over the color of the output using cartoon-like color maps, but this control comes at the cost of a tradeoff between the quality of the generation and the fidelity to the color condition. Exploiting classifier guidance\cite{dhariwal_diffusion_2021} which requires training, in \cite{bansal2023universal}, the authors propose a universal guidance, with a training-free method to control any diffusion models for any condition. In the specific case of color guidance, we noticed that among the existing methods, only a part of the information used for control is preserved when guiding the generation.

We propose a new formulation of the guidance specific to color map control. We rewrite the equations inspired by universal-guidance\cite{bansal2023universal}, and show that, for color maps, contrary to the general case, the scaling of the guidance term should not decrease during diffusion. We call our method \textit{fine color guidance}, we show that we make use of most of the information provided by the color map, without affecting the quality of the generated images.

Since the distortion perception tradeoff highlighted by Blau\textit{.et al}\cite{blau2018perception}, and more particularly at extremely low bitrates \cite{blau2019rethinking}, compression has been taking a step back from the MSE as a reconstruction criterion and focusing on semantic descriptors\cite{pezone2023semantic, bordin2023semantic}. One way of depicting an image is through a high-level color description, giving only a global aspect of what it contains (e.g., a blue spot on the right, green in the middle, blue on the top, etc.). Controlling the color map in the generation of images becomes interesting in that case. Inspired by the framework CoCliCo based on a CLIP and color semantic proposed in \cite{bordin2023coclico}, we propose to apply our fine-color guidance to improve on image compression. We compare the compression framework with another recent generative approach\cite{lei2023text+} and a classical encoder VVC\cite{VVenC} by targeting similar bitrates.

After a brief reminder of how diffusion models work, we present the formulation of our problem in section \ref{sect:Formulation}. We then compare existing color control solutions in section \ref{sect:Diffusion}. We present our proposed method for fine-guidance in section \ref{sect:Fine-guidance} and its application to image compression in an ultra-low bit-rate framework in section \ref{sect:compression}. Finally, in section \ref{sect:evaluation}, we compare our fine color guidance with other existing methods and their impact on compression.

\section{Controlling colors in diffusion models}\label{sect:Formulation}

Our goal is to add control, without requiring training, to any image diffusion model by using color information. Since the latest models work in the latent space of a VAE, control should also be available in these contexts. First, let us recall how diffusion models work.

\subsection{Diffusion models}

\begin{figure}[htbp]
    \centering
    \includegraphics[width=0.4\textwidth]{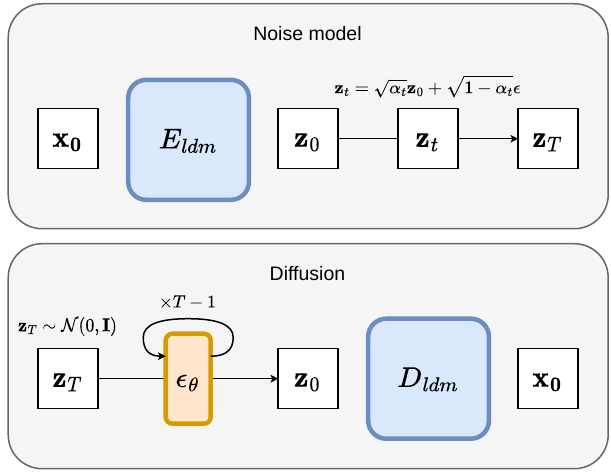}
    \caption{Illustration of the latent diffusion models. Diffusion in the domain of pixels becomes a particular case where there is no encoder or decoders $E_{ldm}=D_{ldm}=I$}
    \label{fig:dm_ldm}
\end{figure}

Diffusion models stand as denoisers, starting from a random gaussian noise, iteratively converging towards a \textit{clean} signal. With the popularity of diffusion models, the need for fast generation and fast training has encouraged the emergence of latent diffusion models\cite{rombach_high-resolution_2022}, diffusing in the latent space of a fixed VAE. In the following, to be more concise, we consider that pixel space diffusion is a specific case of latent diffusion where the VAE is the identity, see Fig. \ref{fig:dm_ldm}.

\subsubsection{Noise model}

Diffusion models learn to invert an iterative noise process on the signal. The model denoises on a series of timesteps $[0,T]$ with scaling noise controlled by the parameters $\{ \alpha_t\}_{t=0}^T$. For a given signal $\bm z_0$ and at a given timestep $t$, the added noise follows the schedule:
\begin{gather}
    \bm z_t = \sqrt{\alpha_t} \bm z_0 + \sqrt{1-\alpha_t} \epsilon \label{eq:z_t}\\
    \text{where \hspace{6pt}}  \epsilon \sim \mathcal{N}(\bm 0,\bm I) \notag
\end{gather}
The mean of $\bm z_t$ is progressively shifting towards $0$ while the amount of noise is increasing with $t$. Going from the value of the signal $\bm z_0$ at $t=0$ to a normal distribution at $t=T$. The diffusion model reverse this noise by estimating the value of $\epsilon$ at each step.\\

\subsubsection{Standard diffusion models}

The diffusion model learns to predict the noise $\epsilon$ added to $\bm z_t$ in \eqref{eq:z_t} from the observation of the pair $(\bm z_t, t)$. The training loss for the timestep $t$ can be written as:
\begin{equation}
    \mathcal{L}_t(\theta) = \mathbb{E}_{q(\bm x_0);\epsilon\sim\mathcal{N}(\bm0, \bm I)} \left[ \left \| \epsilon - \epsilon_\theta(\bm x_t(\bm x_0 ,\epsilon), t) \right \|^2 \right] \label{eq:diff_loss}
\end{equation}
When generating an image with a diffusion model, the noising process is reversed. Starting from a random gaussian noise as $\bm z_T$, we iteratively estimate the next sample $\bm z_{t-1}$ until reaching $t=0$. Different methods -- called schedulers -- can be used to estimate $\bm z_{t-1}$, for instance in denoising diffusion implicit model (DDIM)\cite{song2020denoising}:
\begin{equation}
\resizebox{0.91\hsize}{!}{%
$\bm z_{t-1} = \sqrt{\alpha_{t-1}} \underbrace{\left( \frac{\bm z_t - \sqrt{1-\alpha_t} \epsilon_\theta(\bm z_t,t)}{\sqrt{\alpha_t}} \right)}_\text{"predicted $\bm z_0: \hat{\bm z}_0(\bm z_t,t)$"} + \underbrace{\sqrt{1-\alpha_{t-1}}\epsilon_\theta(\bm z_t,t)}_\text{"direction pointing to $\bm z_t$"}$%
} \label{eq:DDIM}
\end{equation}
We are able to estimate a predicted signal $\hat{\bm z}_0(\bm z_t,t)$ at each step of the generation process, refining the prediction as the process is repeated. Reaching the timestep $0$, we get a clean signal $\bm z_0$, which can de be decoded by the VAE to obtain a generated image $\bm x_0$.

\subsection{Formulation}

Using an already trained and fixed diffusion model $\epsilon_\theta$, that can generate images $\bm x_0(\epsilon_\theta)$. We would like, through the use of a color map $\bm c$, to control the outputs of the model in order such that the generated image has the global appearance of the color map. We also would like that this control should not come at the expense of the generative power of the model, hindering the realism or quality of images. Control should be achieved without retraining, keeping the parameters $\theta$ of the model intact, but modifying the noise estimation, noted as $\tilde {\epsilon}_\theta(\bm c) = \tilde {\epsilon}_\theta(\epsilon_\theta, \bm c)$ to relax the notations. The problem can then be formulated as:
\begin{gather}\label{eq:formulation}
    \min_{\tilde { \epsilon}_\theta} d(\bm c, \bm x_0(\tilde{ \epsilon}_\theta(\bm c)))  \text{  s.t. } \\
    \Psi(\bm x_0(\epsilon_\theta)) = \Psi(\bm x_0(\tilde{ \epsilon}_\theta(\bm c)))  \notag
\end{gather}
where $d$ measures the consistency between a color map and an image, and $\Psi$ measures realism. In other words, the control function should minimize the error of color map while keeping the generative power. Since the control is done during the diffusion process, the estimation of the correction has to be assessed using noisy observation of the latent image $\bm z_t$, this additional level of abstraction makes it difficult to have an explicit function.

\section{Existing solutions}\label{sect:Diffusion}

Several solutions already address the problem previously formulated, however, none of them satisfies all the conditions: no-training, functioning for LDMs and minimizing the error. We present the pros and cons of several methods in our context and motivate the need for a specific method for color guidance, especially in the case of latent diffusion models.

\begin{figure}[htbp]
    \centering
    \begin{subfigure}[b]{0.24\textwidth}
        \centering
        \includegraphics[width=\textwidth]{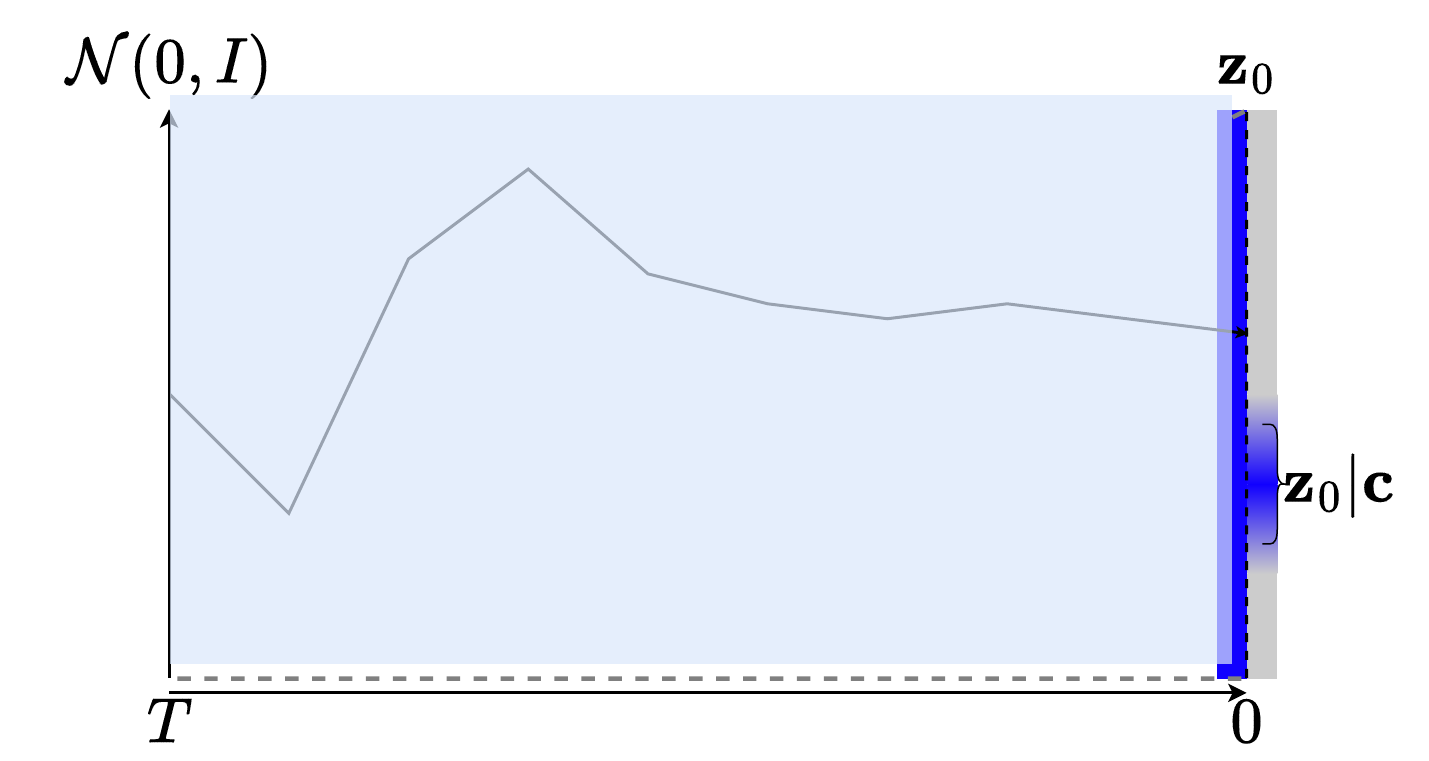}
        \caption{Standard diffusion}
        \label{fig:standard}
    \end{subfigure}
    \begin{subfigure}[b]{0.24\textwidth}
        \centering
        \includegraphics[width=\textwidth]{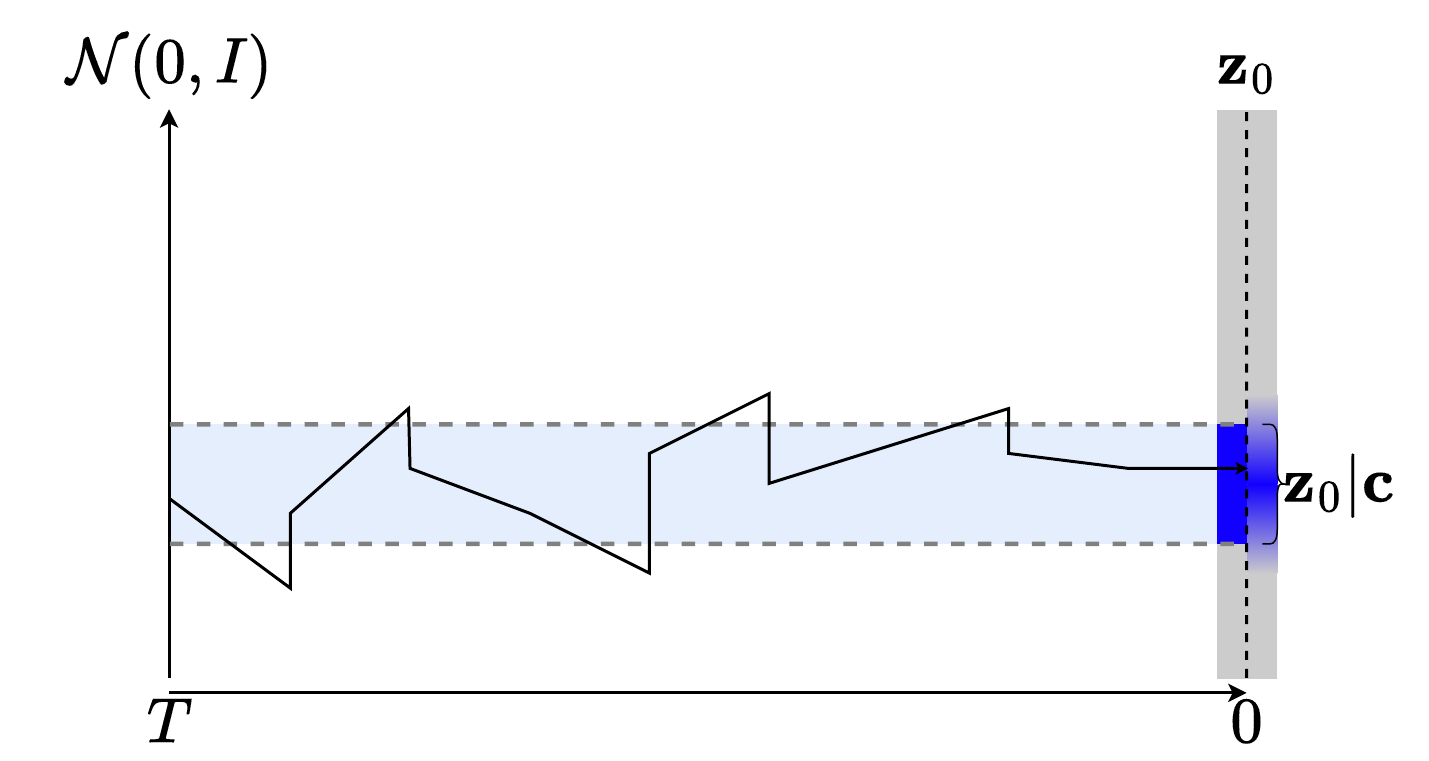}
        \caption{Enforced conditioning}
        \label{fig:enforced}
    \end{subfigure}
    \begin{subfigure}[b]{0.24\textwidth}
        \centering
        \includegraphics[width=\textwidth]{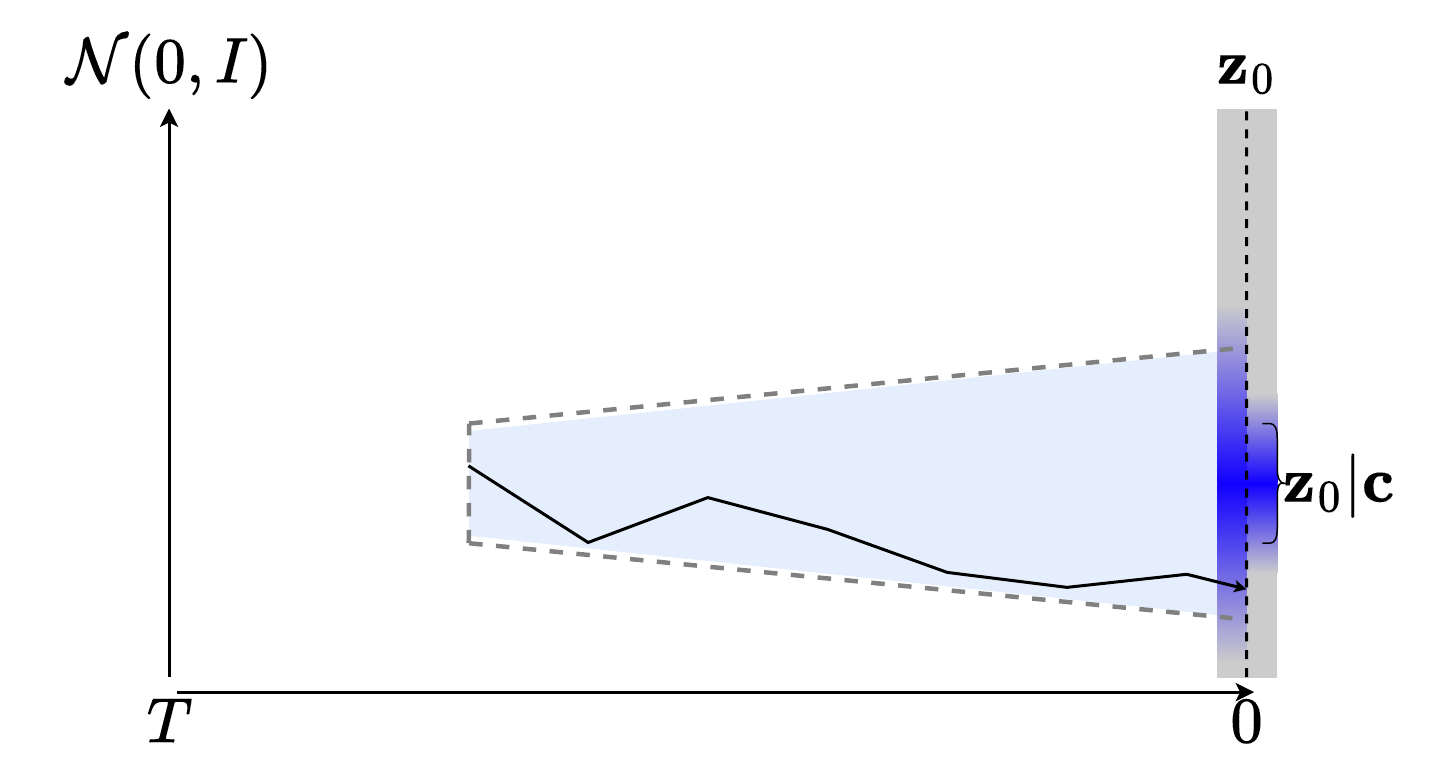}
        \caption{Initialized diffusion}
        \label{fig:initialized}
    \end{subfigure}
    \begin{subfigure}[b]{0.24\textwidth}
        \centering
        \includegraphics[width=\textwidth]{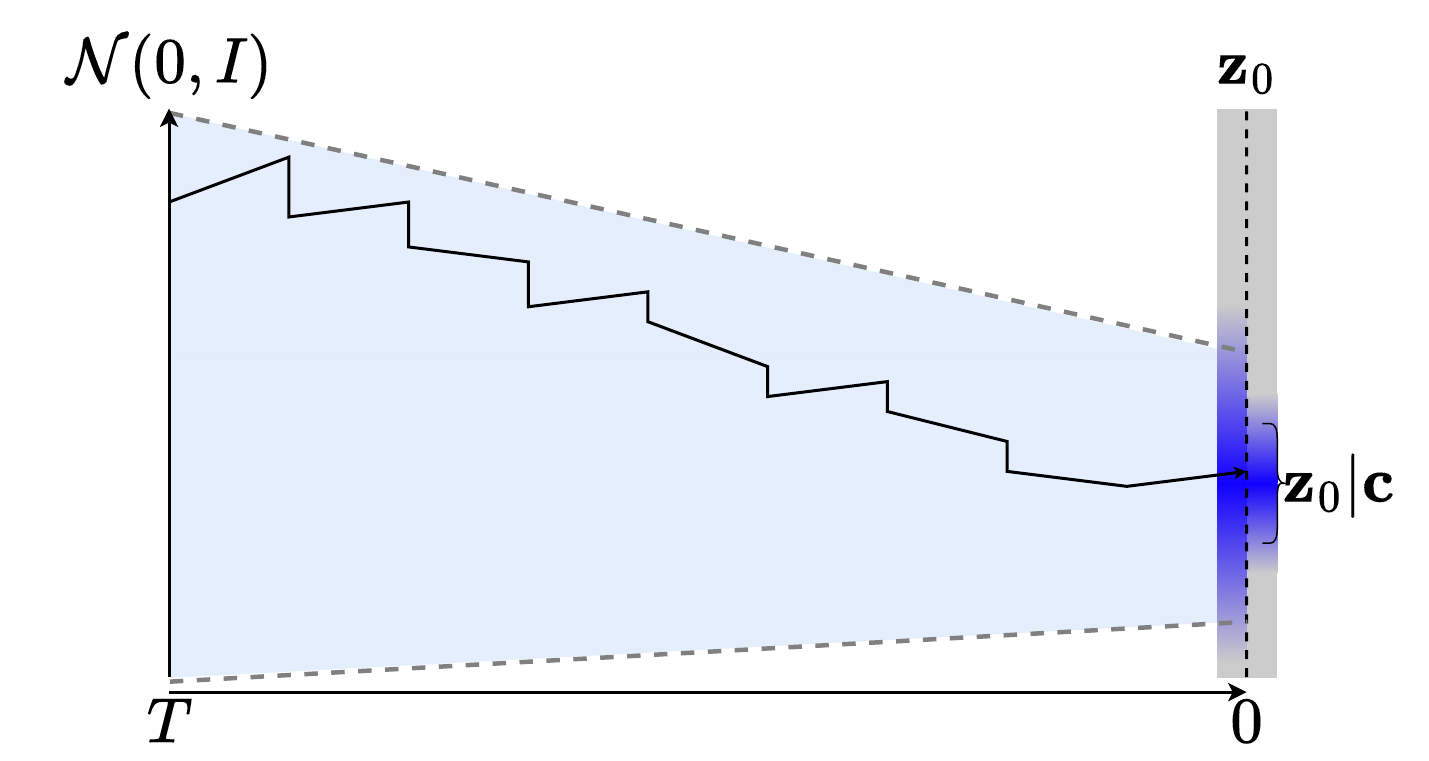}
        \caption{Guided diffusion}
        \label{fig:guided}
    \end{subfigure}
    \caption{Illustration of different method to control the diffusion process. The region of possible sample is illustrated in blue. The black arrow shows a possible diffusion schedule. The first figure shows a standard diffusion process with no control. Our goal is to focus the output region as close as possible of $\bm z_0| \bm c$ as possible.}
\end{figure}

\subsection{Conditional diffusion models}\label{sect:conditional}
A conditional diffusion model follows the same principle as a standard diffusion model, the main difference is that the prediction of the noise is done over $(\bm z_t, t, \bm c)$. The condition, $\bm c$ is then given as input of the diffusion model at each timestep, with parameters $\theta$ trained to take $\bm c$ into account. Conditional models are widely popular for text-to-image generation or image-to-image generation application, however, to get a conditional diffusion model, it requires either retraining a whole network or fine-tuning an already trained one on the desired condition. Which is something we would like to avoid in our framework.

\subsection{Enforcing colors in diffusion model}\label{sect:enforce}
Given a diffusion model $ \epsilon_\theta$, it is sometime possible to enforce a condition on the output of the model, ensuring that the output \textit{exactly} follows our condition. For color map conditioning, it is the case, but only when the diffusion process is done in the pixel space, which means trained to directly generate images from noise without the auto-encoder. Then, for a given image $\bm z_0$, we can separate it as the sum of its lower and higher frequencies uniquely $\bm z_0= l(\bm z_0)+h(\bm z_0)$ for specified functions $l, h$, for instance, authors in \cite{you2023indigo} propose to use a wavelet decomposition. It is then possible to enforce the value of the low (or high frequencies) of the generated image by replacing at each step the generated low frequencies by our target $\bm c$. With $\hat {\bm z}_0(\bm z_t, t)$ the predicted $\bm z_0$ in \eqref{eq:DDIM}, we edit the value at each step, replacing it by:
\begin{equation}
    \hat {\bm z}_0(\bm z_t, t, \bm c) = \bm c + h(\hat {\bm z}_0(\bm z_t, t)) \label{eq:sep}
\end{equation}
At each step, the low frequencies of the sample are the one we enforce, and only the higher frequencies are the one generated. Enforced diffusion is illustrated in Fig. \ref{fig:enforced}, where vertical movement in the arrow indicates the correction of the low frequencies. This method is similar to the inpainting process described in \cite{lugmayr2022repaint}, where a mask is used to do the partitioning of $\bm z_0$ rather than a DCT. This method can be used when the constraint on $\bm c$ needs to be strong and when the diffusion is done in the pixel space. This does not allow a trade-off and to scale the proximity with the condition. Moreover, with the popularity of diffusion models, the speed of their training and sampling time has been a major concern and most of the models are now processing in the latent space of VAE to reduce the dimension, then, the same partitioning is no longer available.

\subsection{Initialized DM}\label{sect:init}

Another method to control the generation of the diffusion model is initialization. First introduced in \cite{meng2021sdedit}, the diffusion process can be initialized with information on the signal instead of using a random noise. With this method, we influence the diffusion process by giving partial information on the signal we desire to get. In practice, it is done by starting the generation at a later timestep $0<\tau<T$ with information on the color degraded at the corresponding scaling of noise:
\begin{equation}
    \bm z_{\tau} = \sqrt{\alpha_{\tau}}E_{ldm}(\bm c) + \sqrt{1-\alpha_{\tau}}  \epsilon \label{eq:init_color}
\end{equation}
Depending on the quantity of information we want to preserve, we can adjust the starting timestep to add more or less noise. This method is quite simple, does not require training, and does even shorten the generation time, however, the generation process is not controlled towards the output and can still drift away from the initialization as illustrated in the diagram Fig. \ref{fig:initialized}. Also, authors in \cite{meng2021sdedit} shows the presence of a tradeoff between the fidelity to the condition $\bm c$ and the quality of the generated signal. This tradeoff depends on the starting timestep $\tau$. Indeed, the higher the timesteps, the more noise we add and the further we get from $\bm c$ and inversely.

\subsection{Guidance for diffusion model}

Instead of initializing as in \ref{sect:init}, or replacing as in \ref{sect:enforce}, authors in \cite{dhariwal_diffusion_2021} propose to guide the diffusion at each iteration towards an image that fits the color map condition. Moreover, in \cite{bansal2023universal} authors propose a method they call universal to guide the output towards a condition without any training.

First introduced as \textit{classifier guidance}, to condition on a set of classes, guidance can be extended to any type of condition. The only requirement is to be able to estimate the value of $\nabla_{\bm z_t} \log p_t(\bm c|\bm z_t)$, the gradient of the log likelihood of the condition $\bm c$ given a noisy observation $\bm z_t$. While conditional models (\ref{sect:conditional}) can learn to estimate this term separately through \textit{classifier-free guidance}\cite{ho2022classifier}, classifier guidance uses a trained classifier to do so. A guided diffusion model process is illustrated in Fig. \ref{fig:guided}, the guidance correction is applied after each timestep, iteratively getting closer to the condition.

The term for guidance comes from the parallel between the diffusion model $\epsilon_\theta$ with energy based models. Indeed, if $\bm f_\theta$ is an energy based model trained to predict the log likelihood $\log p(\bm x)$, the corresponding training loss is formulated as:
\begin{equation}
    \mathcal{L}_\sigma(\theta) = \mathbb{E}_{\bm x\sim q(\bm x)\epsilon\mathcal{N}(\bm 0, \bm I)} \left[ \left \| \frac{\epsilon}{\sigma} + \nabla_{\bm x} \bm f_\theta(\bm x + \sigma \epsilon) \right\|^2 \right] \label{eq:ebm_loss}
\end{equation}
Putting the training losses of $\epsilon_\theta$ \eqref{eq:diff_loss} and $f_\theta$ \eqref{eq:ebm_loss} in parallel, we can conclude that:
\begin{equation}
    \epsilon_\theta(\bm z_t ,t) = - \sqrt{1-\alpha_t} \nabla_{\bm z_t} \log p_t(\bm z_t) \label{eq:ebm_parallel}    
\end{equation}
And using the Bayes formula:
\begin{equation}
\nabla_{\bm z} \log(p(\bm z|\bm c)) = \nabla_{\bm z} \log(p(\bm z)) - \nabla_{\bm z} \log(p(\bm c|\bm z)) \label{eq:bayes}
\end{equation}
From which we can then get the classifier guidance formula \cite{song2020score} to control the generation:
\begin{equation}
    \tilde \epsilon_\theta(\bm z_t, t, \bm c) =  \epsilon_\theta(\bm z_t, t) - \sqrt{1-\alpha_t} \nabla_{\bm z_t} \log p_t(\bm c | \bm z_t) \label{eq:classifier_guidance}
\end{equation}
The impact of the guidance is often scaled by a scaling factor $s$, $s\cdot\sqrt{1-\alpha_t} \nabla_{\bm z_t} \log p_t(\bm c | \bm z_t)$. Increasing the scaling factor then focuses the outputs more towards fidelity to the condition, at the cost of the diversity of generation. To simplify and shorten notations, we will note $G(\cdot , \cdot, \cdot)$ the guidance correction operator:
\begin{equation}
    G(\bm z_t, t, \bm c) = - \sqrt{1-\alpha_t} \nabla_{\bm z_t} \log p_t(\bm c | \bm z_t) \label{eq:G}
\end{equation}

Authors in \cite{kwon2022diffusion} propose to measure the log likelihood using the predicted signal $\hat {\bm z}_0(\bm z_t, t)$ \eqref{eq:DDIM} rather than the noisy signal $\bm z_t$. In \cite{bansal2023universal} the authors propose to go further and to approximate the log likelihood $\log p_t(\bm c | \hat {\bm z}_0(\bm z_t, t))$ by a distance between the estimated condition on the predicted signal $\hat {\bm c}_t = \bm c(\hat {\bm z}_0(\bm z_t, t))$, and the given condition. In the case of color guidance, with their formulation, the correction becomes:
\begin{equation}
    G(\bm z_t,t, \bm c) = s \sqrt{1-\alpha_t} \nabla_{\bm z_t} \| \bm c - \hat {\bm c}_t\|^2 \label{eq:univ} 
\end{equation}
where the scale $s$ is specific to colors and needs to be determined empirically. Their approximation comes from the intuition that if the estimated condition is close to the given condition, then the probability that the signal follows this condition should be high. However, this formulation highly varies with the type of condition we give, and obtaining the right value for $\lambda$ is detrimental to guide the model.

\section{Proposed fine color guidance for diffusion models}\label{sect:Fine-guidance}

Inspired by the formulation of universal guidance\eqref{eq:univ}, we propose to study how the estimation of guidance could be formulated in the specific context of color conditioning. We first model the color map before unrolling the formula for guidance.

\subsection{Modeling the color map}\label{sect:color_map}

The color map $\bm c$ represents coarse information on the color of an image $\bm x$. For our problem, we define the color map as the low frequencies of the image. Using the 2d-DCT, $\bm x = \bm V \bm D \bm V^T$, we extract the $m$ lowest frequencies in the DCT. In other words: 
\begin{equation}
\bm c= \bm A \bm x \label{eq:color}
\end{equation}
is a $m$ dimension vector. With $\bm A$ a being thresholding operator on the frequencies in the DCT basis, which can be written as:
\begin{equation}
    \bm A= \bm V_{[m,n]}^g \bm I_m \bm V^T_{[m,n]} \label{eq:color_map}
\end{equation}
where $\bm V_{[m,n]}$ stands for the sub-matrix containing the $m$ first rows and $n$ first columns of $\bm V$, and where $\bm V_{[m,n]}^g$ is the generalized inverse of $\bm V_{[n,m]}$. In the following, we consider that $\bm c$ is a fixed color map of size $m$.

\subsection{Formulation}

The formulation, previously defined in section \ref{sect:Formulation} in equation \eqref{eq:formulation}, can now be adapted and simplified. With the notation of guidance defined in \eqref{eq:classifier_guidance} and \eqref{eq:G}, the formulation becomes:
\begin{gather}
    \min_G ||\bm c - \bm A\bm x_0(\tilde{\epsilon}_\theta(\bm c))||_2^2  \text{  s.t. } \\
    \Psi(\bm x_0(\epsilon_\theta)) = \Psi(\bm x_0(\tilde{ \epsilon}_\theta(\bm c)))  \notag
\end{gather}
we measure the distance between the conditioning color map and the color map of the output with a simple $L²$ norm. We add a guidance term $G$ to the diffusion model to modify the noise estimation $\tilde{\epsilon}_\theta$. The problem lies in writing the explicit formulation of the guidance function $G$ optimally.

\subsection{Unrolling classifier guidance equations}

The guidance of a diffusion model relies on the classifier guidance expression in \eqref{eq:classifier_guidance}. As proposed in \cite{kwon2022diffusion}, we use the predicted signal for the estimation in \eqref{eq:DDIM} rather than the noisy signal.\\

\subsubsection{The predicted signal} $\hat {\bm z}_0(\bm z_t,t)$\\
We know from \eqref{eq:z_t} that for a given sample $\bm z_0$, the noisy samples $\bm z_t$ at timestep $t$ follow a multivariate normal distribution:
\begin{equation}
    \bm z_t \sim \mathcal{N}\left(\sqrt{\alpha_t} \bm z_0, (1-\alpha_t) {\bm I}\right) \label{eq:N_zt}
\end{equation}
Moreover, at any timestep $t$, we can build a prediction $\hat {\bm z}_0$ of the value of $\bm z_0$. Using the expression in \eqref{eq:DDIM}:
\begin{equation}
    \hat {\bm z}_0(\bm z_t,t) = \frac{\bm z_{t} - \sqrt{1-\alpha_{t}} \epsilon_\theta(\bm z_t,t)}  {\sqrt{\alpha_{t}}} \label{eq:N_z0}
\end{equation}
If the diffusion model was ideal, then $\epsilon_\theta(\bm z_t, t)$ would predict exactly the noise added $\epsilon$ and we would have $\hat {\bm z}_0(\bm z_t,t) = \bm z_0$. In practice, however, noisy signals can overlap for different images, especially for higher value of $t$ with a complete overlap at $T$. The model will not be able to correctly predict $\epsilon$ because the solution is not unique. We can thus consider that the prediction error $\Delta \epsilon_t$ is non-zero. We can then rewrite $\hat {\bm z}_0(\bm z_t, t)$:
\begin{align}
    \hat {\bm z}_0(\bm z_t,t) &= \frac{\sqrt{\alpha_t}\bm z_0 - \sqrt{1-\alpha_{t}} (\epsilon_\theta(\bm z_t,t) -  \epsilon)}{\sqrt{\alpha_{t}}} \notag \\
                                &= \bm z_0 - \frac{\sqrt{1-\alpha_t}}{\sqrt{\alpha_t}} \Delta \epsilon_t \label{eq:tilde_z0}
\end{align}

We measured the distribution of the error for a given diffusion model, and we observed that depending on the value of $t$:
\begin{equation}
    \Delta \epsilon_t \sim \mathcal{N}(0, \overline{\lambda}_t^2 {\bm I}) \label{eq:eps_error}
\end{equation}
with $\overline{\lambda}_t$ measured empirically over a set of images, and probably specific to the model. Thus, the predicted sample $\hat {\bm z}_0(\bm z_t,t)$ follows a gaussian distribution centered on the real value $\bm z_0$:
\begin{equation}
    \hat {\bm z}_0(\bm z_t, t) \sim \mathcal{N}\left(\bm z_0, \overline \lambda_t^2\frac{1-\alpha_t}{\alpha_t}{\bf I}\right) \label{eq:N_tilde_z0}
\end{equation}
We now have the distribution of the predicted signal given the noisy observation. Let us now move on to the estimation of the density probability: $p_t(\bm c, \hat {\bm z}_0(\bm z_t,t))$. 

\subsection{Guidance for pixel space diffusion}

In the pixel space (i.e $\bm z=E_{ldm}(\bm x)=\bm x$), at the timestep $t$, we can evaluate $\hat {\bm c}_t$ the estimated color map over the predicted image $\hat {\bm x}_0(\bm x_t,t)$, in other words, the predicted color map:
\begin{align}
    \hat {\bm c}_t(\bm x_t,t) &= \bm A \hat {\bm x}_0(\bm x_t, t) \notag\\
                              &= \bm c + \overline{\lambda}_t\frac{ \sqrt{1-\alpha_t}}{\sqrt{\alpha_t}} \bm A \epsilon \label{eq:tilde_ct}
\end{align}
The distribution of the estimated color map is then also a gaussian centered on the targeted color map $\bm c$. Which can also be written as:
\begin{equation}
    \bm c|\bm x_t \sim \mathcal{N} \left(\hat {\bm c}_t, \overline{\lambda}_t^2\frac{1-\alpha_t}{\alpha_t} {\bm I_m}\right) \label{eq:N_tilde_ct}
\end{equation}

We now have an explicit formula of the distribution of $\bm c|\bm x_t$. We also get the value of $p_t(\bm c|\bm x_t)$, the probability density of a multivariate distribution, that we were looking for:
\begin{equation}\label{eq:pt_c}
p_t(\bm c|\bm x_t) = \frac{1}{\sqrt{(2\pi)^{m}}}\exp\left(-\frac{1}{2\overline{\lambda}_t}\frac{\sqrt{\alpha_t}}{\sqrt{1-\alpha_t}} \| \bm c - \hat {\bm c}_t\|^2_2\right)
\end{equation}

We can then use \eqref{eq:pt_c} in the guidance term in \eqref{eq:G}. The guidance correction term for a color map in a pixel space diffusion model then becomes:
\begin{equation}\label{eq:guidance_dm}
    G(\bm x_t,t, \bm c) = \frac{\sqrt{\alpha_t}}{2\overline{\lambda}_t} \nabla_{\bm x_t} \|\bm c- \hat {\bm c}_t\|^2_2
\end{equation}
We can see that the correction term to guide the color in the diffusion model is similar to the one proposed in universal guidance \eqref{eq:univ}, as both uses a $L^2$ norm. Though, there are still substantial differences. Firstly, the coefficient in front of the norm scales differently with $t$. Indeed, in \eqref{eq:univ}, $\sqrt{1-\alpha_t}$ decreases, guiding less as the diffusion progresses, whereas we observed that $\frac{\sqrt{\alpha_t}}{2\overline{\lambda}_t}$ preserve the importance of guidance along the process. Moreover, once $\overline{\lambda}_t$ is determined, there is no need for tuning any parameter in the loss. Guidance in pixel space brings fewer guarantees than an enforced approach such as \cite{you2023indigo}. However, since it takes less advantage of the image in pixel space, it can be adapted to latent diffusion models and be applied to the state-of-the-art models.

\subsection{Guidance for latent diffusion}
While in the pixel space we had access to an estimated image, here, we only have an estimated latent vector. However, we can still approximate an image as $\hat {\bm x}_0 = D_{ldm}(\hat {\bm z}_0)$, the decoded predicted latent. Similarly to what we had in the pixel space, we would like to know the distribution of the predicted images depending on the timestep. To do so, since the distribution of $\hat {\bm z}_0(\bm z_t,t)$ at each timestep is a multivariate normal, we look at the response of the decoder $D_{ldm}$ when gaussian noise is added to the latent vector. In other words, we observe the behavior of the decoder to $\bm z_\lambda \sim \mathcal{N}(\bm z, \lambda I)$ for a wide range of $\lambda$ and images $\bm z$. We found that for small increase in noise, the decoded image is not far from the original one, while still recovering gaussian noise at the output. Moreover, we model the amount of noise after the decoding as a linear function of $\lambda$, estimating the parameters $\overline a, \overline b$ such as:
\begin{equation}
    D_{ldm}(\bm z_\lambda) \sim \mathcal{N}\left(D_{ldm}(\bm z) + \overline a \lambda, \overline b^2 \lambda^2 \bm I\right)
\end{equation}
Adding noise in the latent space thus provokes a shift of the mean in the decoded image. The decoded image distribution becomes:
\begin{equation}
    \hat {\bm x}_0(\bm x_t,t) \sim \mathcal{N}\left(\bm x_0 + \overline{a}\overline{\lambda}_t\frac{ \sqrt{1-\alpha_t}}{\sqrt{\alpha_t}}, \overline{b}^2\overline{\lambda}_t^2\frac{1-\alpha_t}{\alpha_t} I \right)
\end{equation}

\begin{figure}
    \centering
    \includegraphics[width=0.4\textwidth]{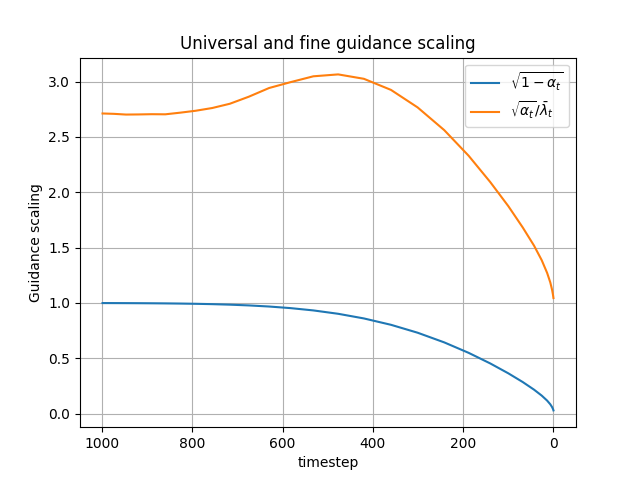}
    \caption{Scaling of the gradient in guidance. In universal-guidance, the scaling decreases to $0$, we observed that theoretically, guidance for color should follow another guidance schedule.}
    \label{fig:guidance_scale}
\end{figure}

From here, the computations are similar to what we had in the pixel space, just taking into account the shift of the mean, the predicted color map becomes:
\begin{align}
    \hat {\bm c}_t(\bm x_t,t) &= \bm A \hat {\bm x}_0(\bm x_t, t) \notag\\
                                &= \bm c + \overline{a}\overline{\lambda}_t \frac{\sqrt{1-\alpha_t}}{\sqrt{\alpha_t}} \bm A \mathbbm 1_{hw} + \overline{b}\overline{\lambda}_t\frac{ \sqrt{1-\alpha_t}}{\sqrt{\alpha_t}} \bm A \epsilon
\end{align}
and the distribution of the estimated color map becomes:
\begin{equation}
    \hat {\bm c}_t(\bm x_t,t) \sim \mathcal{N} \left(\bm c + \overline{a}\overline{\lambda}_t\frac{\sqrt{1-\alpha_t}}{\sqrt{\alpha_t}} \bm A \mathbbm 1_{hw}, \overline{b}\overline{\lambda}_t\frac{\sqrt{1-\alpha_t}}{\sqrt{\alpha_t}} {\bf I}\right)
\end{equation}
The guidance correction term using a color map in a LDM then is:
\begin{equation}\label{eq:guidance_ldm}
    \resizebox{0.89\hsize}{!}{%
    \boxed{G(\bm z_t,t, \bm c) = \frac{\sqrt{\alpha_t}}{2\overline{b}\overline{\lambda_t}} \nabla_{\bm z_t} \|\bm c -  \hat {\bm c}_t - \overline{a}\frac{\overline{\lambda}_t \sqrt{1-\alpha_t}}{\sqrt{\alpha_t}} \bm A \mathbbm 1_{hw}\|^2}
    }
\end{equation}
The guidance correction term is similar to the pixel space \eqref{eq:guidance_dm}. The scaling is slightly different, and we take into account the shifting of the mean produced by the decoder. For an ideal pair of encoder and decoder, we then have $(\bar a, \bar b) = (0, 1)$, which results in the pixel space guidance. When compared to universal guidance \eqref{eq:univ} we can see that the scaling coefficient in front of the gradient still differs from the one they proposed. The value of the guidance should not decrease with the value of $t$ in $\sqrt{1-\alpha_t}$ but rather follow the ratio $\sqrt{\alpha_t}/\overline{\lambda_t}$ which remains high along $t$ for the diffusion model observed. We present, in Fig. \ref{fig:guidance_scale}, the comparison between the two guidance schedule, contrary to universal-guidance, the value does not reach $0$ towards the last steps. In other words, the theory states that one should guide the generation with the color along the diffusion process and not only at the start. We note that this observation is specific to our formulation of the color maps, but could also be generalized to other conditions computed as a matrix product.

\section{Application to compression}\label{sect:compression}

The most recent diffusion models can now generate a high variety of images with impressive quality \cite{betker2023improving}. When applied to generative compression, they can either be used at high bitrates to generate small details and avoid artifacts \cite{agustsson2023multi}, and at lower bitrates, they can be used as decoder using semantic information for conditional generation \cite{pan2022extreme, bordin2023semantic, grassucci2023generative}. Our definition of color map \ref{sect:color_map}, particularly aligns with image compression purposes. Here, we propose to follow the pipeline of CoCliCo\cite{bordin2023coclico} and we represent an image as a combination of semantic and color information. However, to best harness the information on the color that is given, we propose to rely on our formulation of guidance to condition the generation.

\subsection{Compression framework}

The image is encoded into a vector pair $\sigma = (\sigma_s, \sigma_c)$ of semantic and color information, it is decoded using a generative decoder conditioned on the semantic, and guided to correspond to the color. The goal of the compression becomes to generate images with high realism while preserving the maximum semantic information contained in $\sigma$ at a given, extremely low, bitrate.

\begin{figure*}[htbp]
    \centering
    \includegraphics[width=0.8\textwidth]{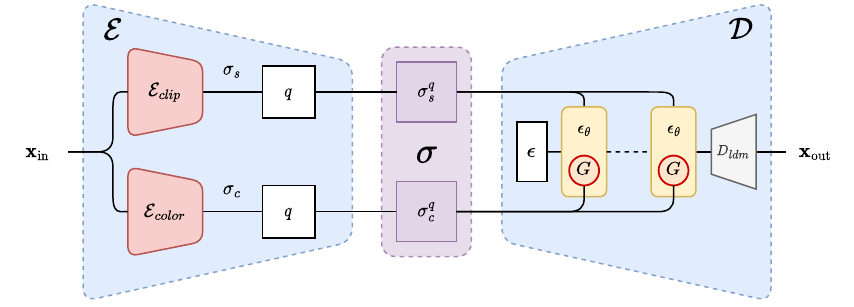}
    \caption{Encoding-decoding pipeline of our framework. Our guidance term $G$ is added over the CoCliCo framework based on CLIP and color semantic. We provide a color correction during generation.}
    \label{fig:app_compression}
\end{figure*}

We call $\Phi$ the semantic projection of an image. The range of the function $\Phi$ then defines a semantic space. We measure the loss of semantic as a distance $d$ in this semantic space. Using this definition, noting $\mathcal{E},\mathcal{D}$ the encoder and decoder ${\bm x}_{\rm{out}}=\mathcal{D}(\mathcal{E}(\bm x_{\rm{in}}))$, we can then formulate the problem of image compression as:
\begin{gather}\label{eq:form_comp}
    \min_\mathcal{D} d(\Phi(\bm x_{\rm{in}}), \Phi(\bm x_{\rm{out}}))  \text{  s.t. } \\
    \Psi(\bm x_{\rm{out}})< \Psi_{min}  \text{ and } R(\mathcal{E}(\bm x_{\rm{in}})) < R_{min} \notag
\end{gather}
where $R$ is the function of the rate and $\Psi$ measures the realism of an image. Similarly to the CoCliCo\cite{bordin2023coclico} framework, we want to minimize the loss of semantic while preserving high realism at a given rate. In our context, the decoder is a fixed function given by the diffusion model that can be influenced by the guide $G$. The rate is measured on the encoded image, i.e, on the color map and the semantic. Seeing the general color aspect as semantic information, the distance $d$ defined here in \eqref{eq:form_comp} encompasses the color metric defined in section \ref{sect:Formulation} in \eqref{eq:formulation}.

\subsection{Representation of the semantic}

The image is encoded in two parts: the semantic $\sigma_s$ and the color $\sigma_c$. We already described the choice and motivation for the color map, but we haven't discussed the semantic yet. As proposed in the CoCliCo framework \cite{bordin2023coclico}, we choose to use the foundation model CLIP\cite{radford2021learning} to encode the semantic of an image. This choice is motivated firstly by the capacity of abstraction of this model but also by its popularity. Indeed, as CLIP is used to encode images and text in the same projection space, this model has been an advantageous choice for text representation in recent text-to-image models, which lead to a large amount of available trained models.

In Fig. \ref{fig:app_compression}, we illustrate the proposed encoding-decoding pipeline. On one hand, $\sigma_s$ is extracted using CLIP as a semantic encoder $\mathcal{E}_s$, and on the other the color map $\sigma_c$ is computed using frequency decomposition. Both the CLIP vector and the color map are quantified as illustrated in Fig. \ref{fig:app_compression} to reduce the rate. The color map is encoded as an $m\times m$ image in YUV 4:2:0 format with a uniform color palette over $b_c$ bits per channel. The CLIP latent vector is clamped then uniformly quantized on $b_s$ bits per dimension as:
\begin{align}\label{eq:quantize}
    \sigma_c^q = \lfloor \frac{\sigma_{c}}{b_s} \rfloor q + \frac{b_s}{2}
\end{align}
With a clip vector of dimension $768$, the rate of an image then becomes:
\begin{align}\label{eq:rate}
    R(\sigma^q) &= R(\sigma_{c}^q) + R(\sigma_s^q) \notag \\
                   &= - \log_2(b_s) * 768 + 1.5 * b_{c} * m^2
\end{align}
where the 1.5 coefficient comes from the YUV 4:2:0 format.

\subsection{Decoding the semantic with a guided diffusion model}

The diffusion model used for decoding thus takes two conditions as inputs: a learned semantic conditioning on $\sigma_s$, and a guided color conditioning on $\sigma_c$. The conditional diffusion model $\epsilon_\theta$ predicts the noise $\epsilon$ from the triplet $(\bm z_t, t, \sigma_s)$ as described in section \ref{sect:conditional}. Our contribution lies in the correction term using the color map guidance term \eqref{eq:guidance_dm} or \eqref{eq:guidance_ldm} respectively for image and latent diffusion during the generative decoding of images. We then have the following the noise estimation function, based on a CLIP model:
\begin{equation}\label{eq:double_guidance}
    \tilde { \epsilon}_\theta(\bm z_t, t, \sigma_s, \sigma_c) = \epsilon_\theta(\bm z_t, t, \sigma_s) + G(\bm z_t, t, \sigma_c)
\end{equation}

\section{Evaluation}\label{sect:evaluation}

We compare in this section several methods to control colors, previously described. We show that our guidance can be applied to pixel space diffusion, but more importantly, it also performs well when applied to latent diffusion models, outperforming current available methods on the latter. We then apply our formulation of guidance in the context of compression in the CoCliCo scheme and show how it compares to other extremely low-bitrates and semantic focused methods.

\subsection{Performance of color control}

\subsubsection{Guidance in the pixel space}

\begin{figure}
    \centering
    \includegraphics[width=0.4\textwidth]{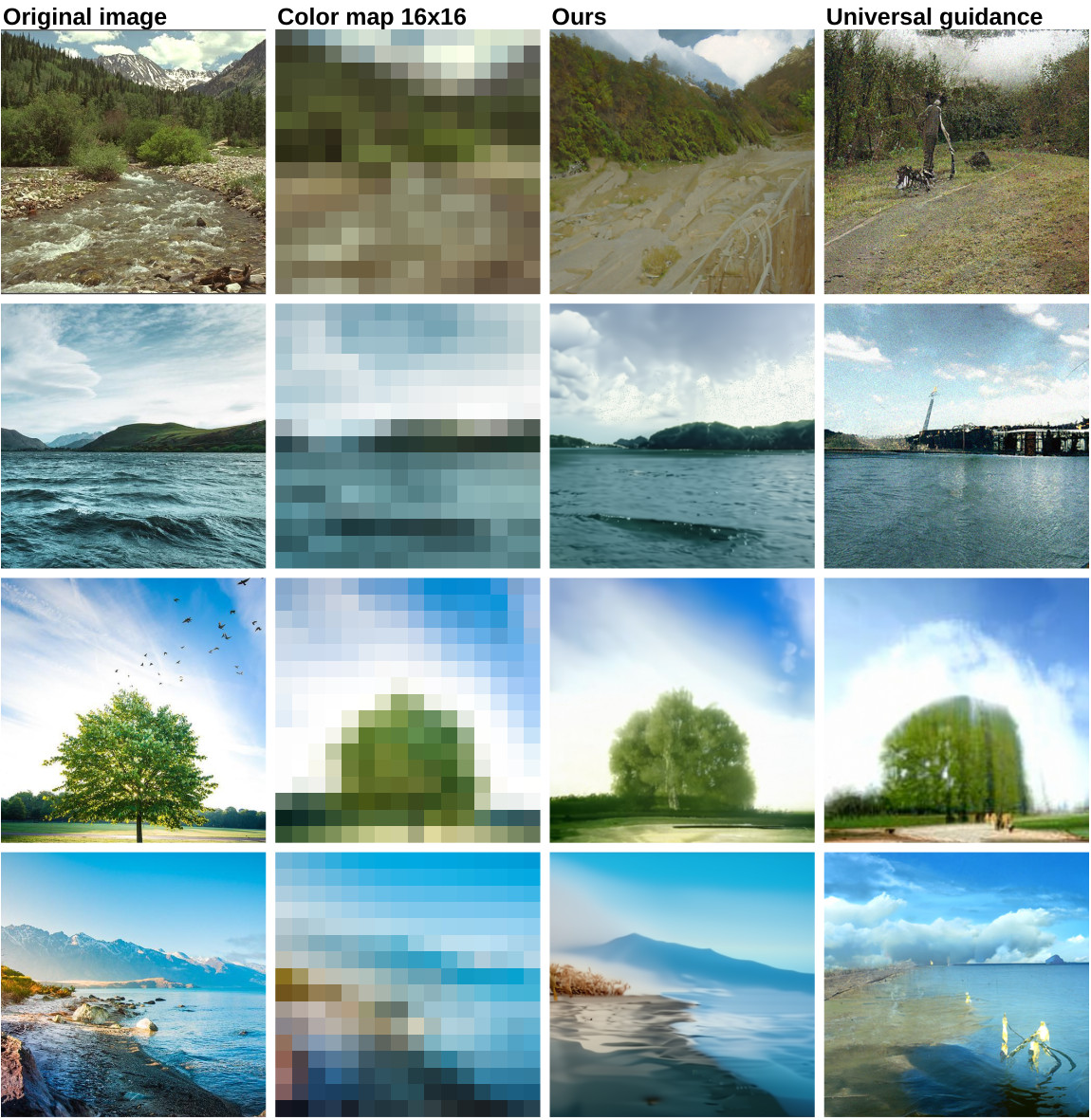}
    \caption{Color guidance applied to pixel space diffusion models.}
    \label{fig:guidance_dm}
\end{figure}

We showcase the guidance for pixel space diffusion. We use an unconditional diffusion model presented in \cite{dhariwal_diffusion_2021} and pretrained for the generation of $512\times512$ images. This model does not use semantic information on the image and does not always produce realistic images, but is sufficient to appreciate the impact of color guidance. We show in Fig. \ref{fig:guidance_dm} some images generated using our guidance terms presented in equation \eqref{eq:guidance_dm} and universal guidance \eqref{eq:univ}. We can see that the generated images follow the given color maps, with more accuracy than when using universal guidance.

\begin{figure*}[htbp]
\centering
    \includegraphics[width=0.95\textwidth]{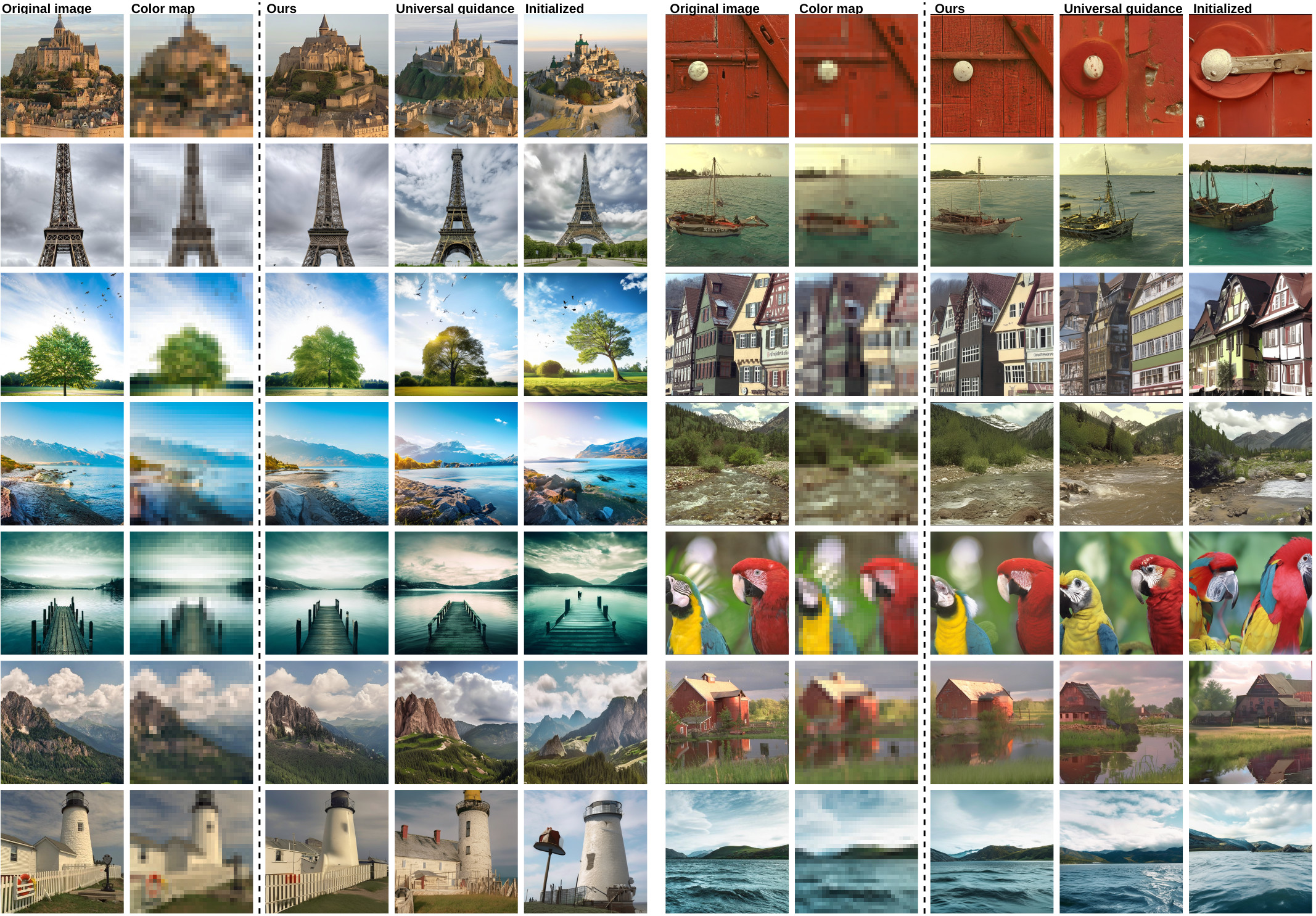}
    \caption{Visual comparison between the different methods to control colors.}
   \label{fig:color_control}
\end{figure*}

\begin{figure*}[htbp]
    \centering
    \includegraphics[width=0.85\textwidth]{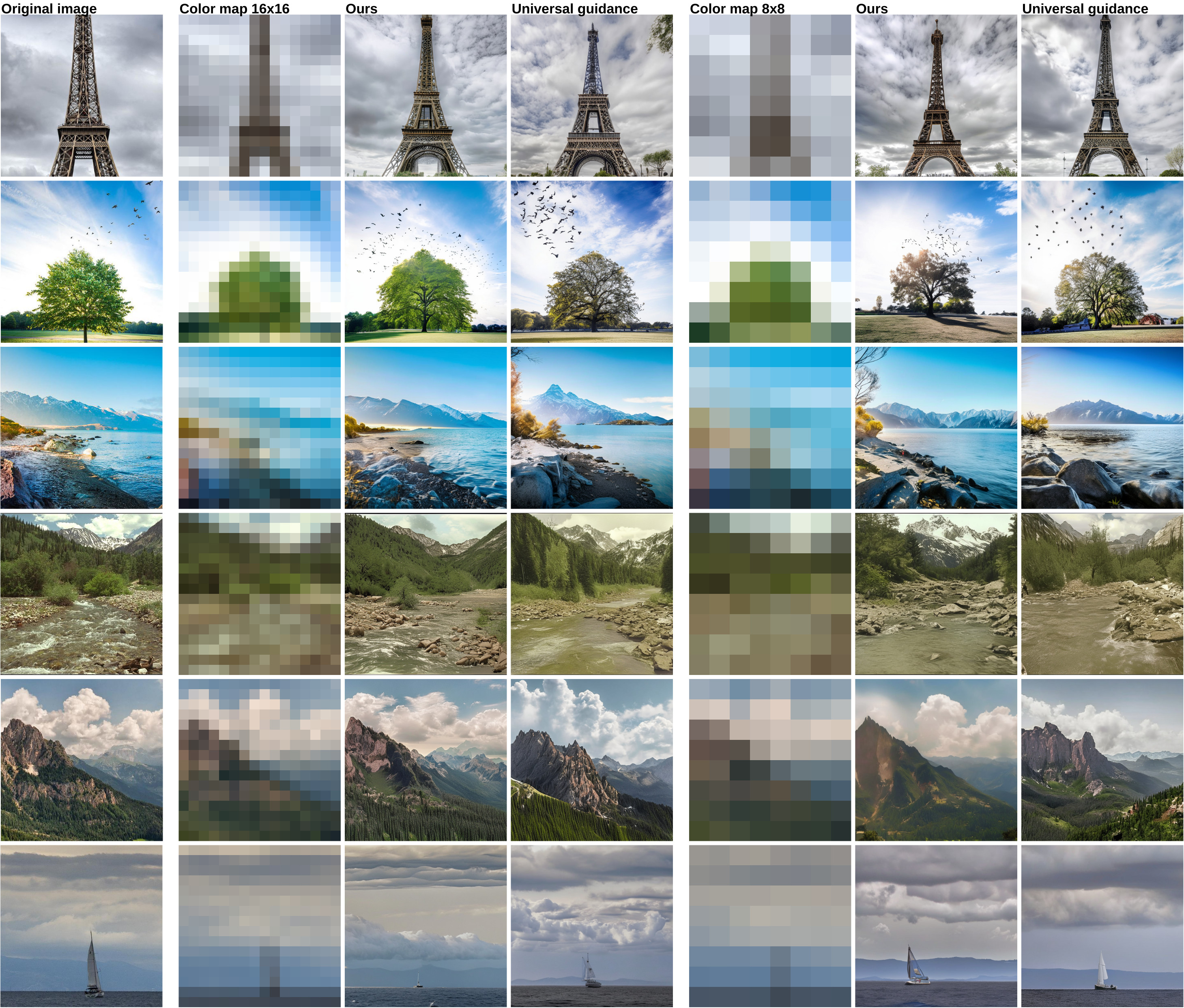}
    \caption{Color map guidance at different resolutions.}
    \label{fig:other_res}
\end{figure*}
\begin{figure*}[htbp]
    \centering
    \includegraphics[width=0.75\textwidth]{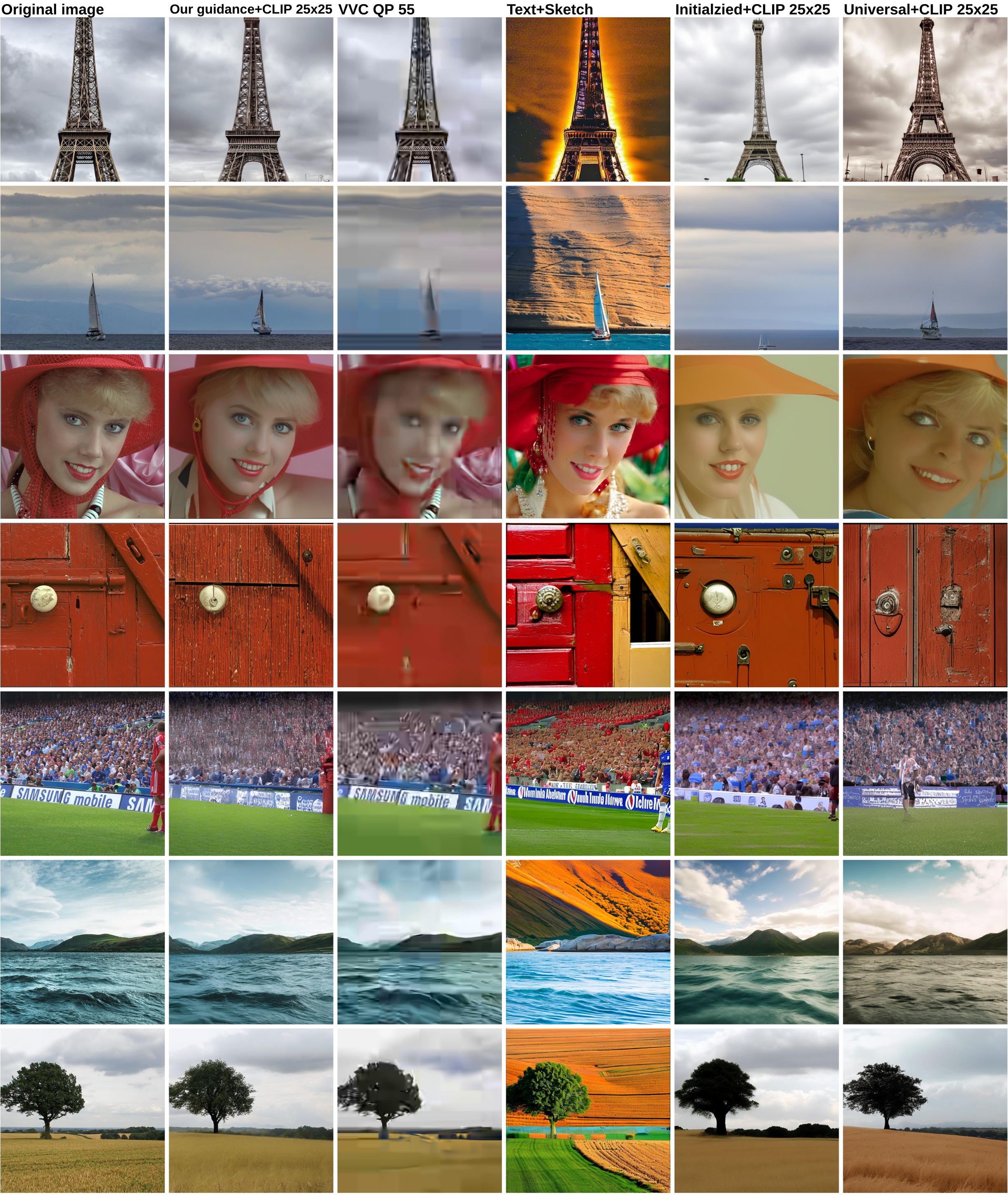}
    \caption{Compression with $25\times25$ resolution color maps.}
    \label{fig:compression25}
\end{figure*}
\begin{figure*}[htbp]
    \centering
    \includegraphics[width=0.8\textwidth]{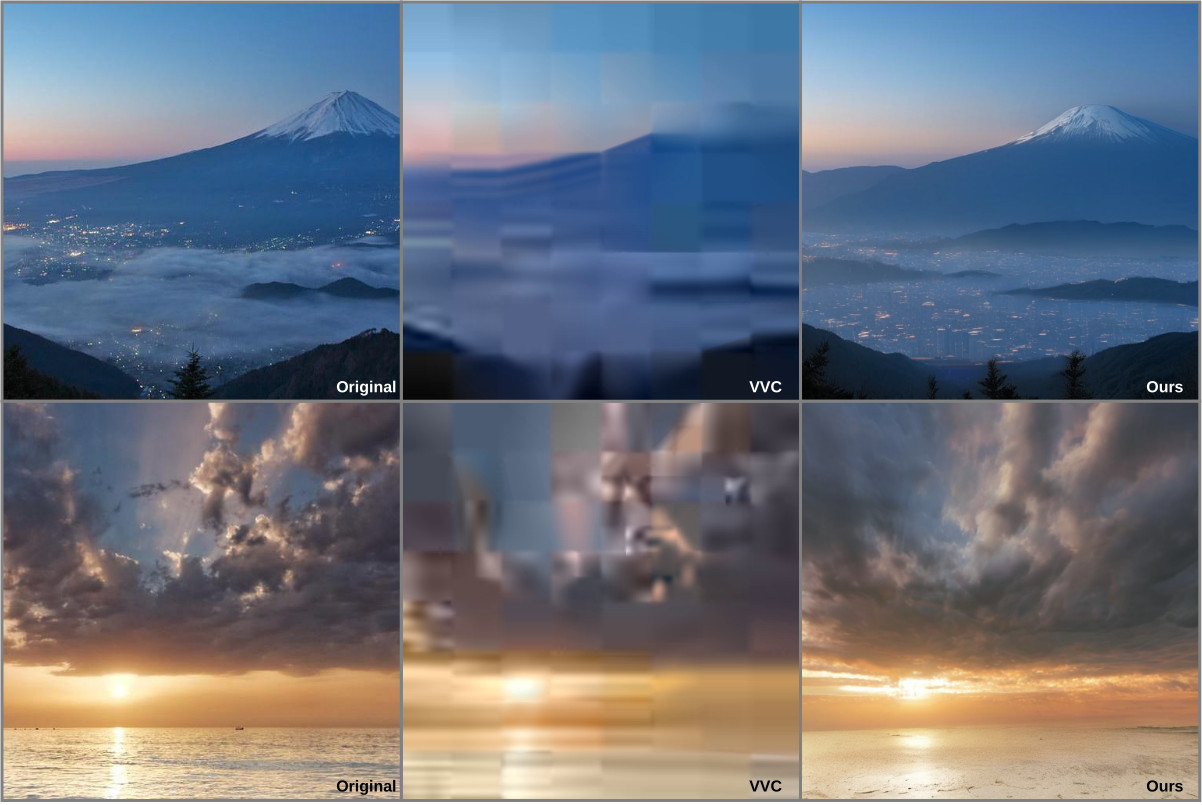}
    \caption{Images decoded using 25x25 resolution color maps and quantized CLIP latent for encoding.}
    \label{fig:big_compression}
\end{figure*}
\begin{figure*}[htbp]
    \centering
    \includegraphics[width=0.7\textwidth]{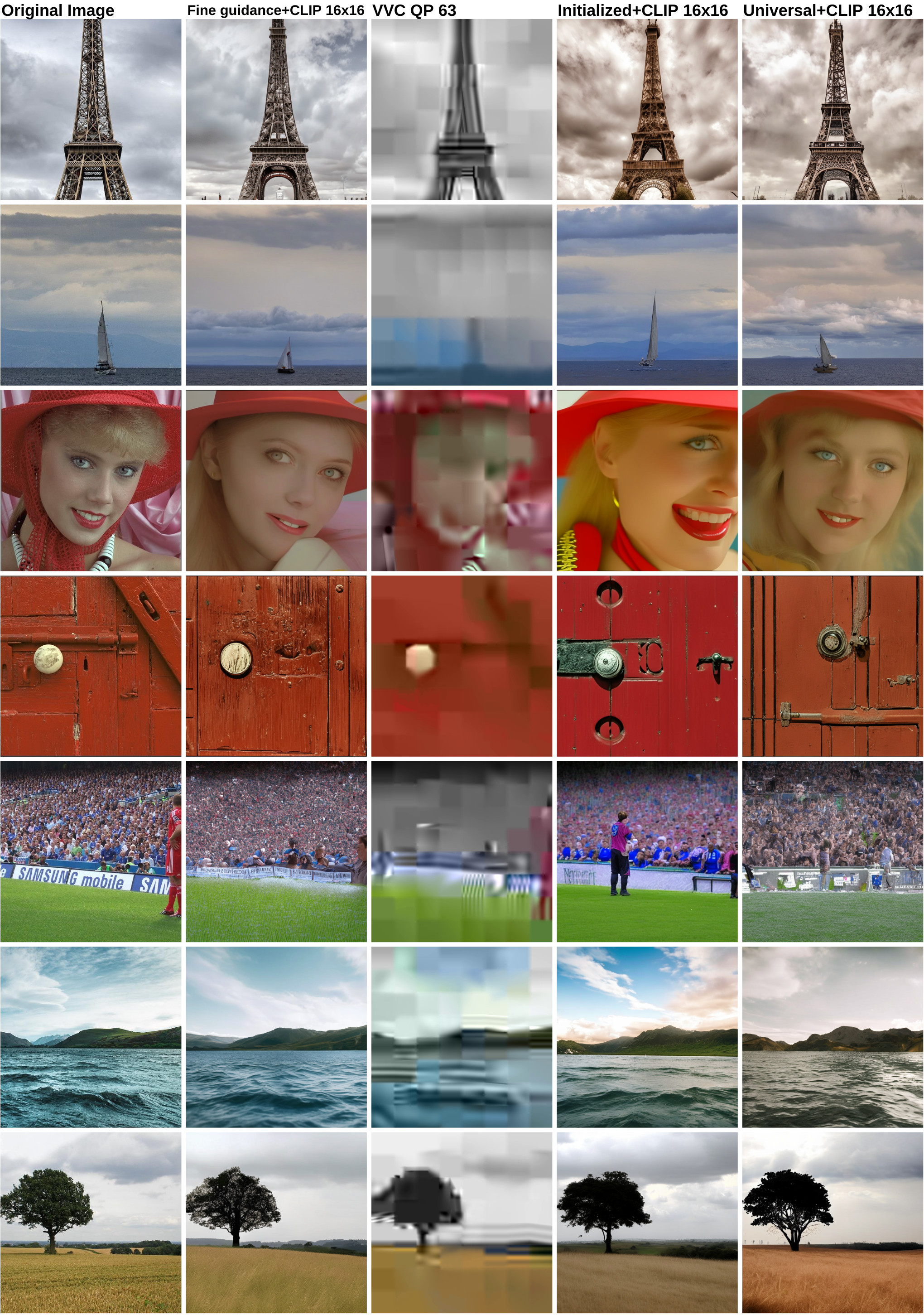}
    \caption{Compression with $16\times16$ resolution color maps.}
    \label{fig:compression16}
\end{figure*}
\subsubsection{Guidance in the latent space}

For latent diffusion guidance, we use another model to compare the different methods to control colors. The model we use is a LDM conditioned on CLIP latent vectors as inputs for text-to-image synthesis. We repurposed it for image-to-image synthesis using the CLIP image encoder rather than the text encoder which refers to $\mathcal E_{clip}$ in Fig. \ref{fig:app_compression}. The weights of the model are available on diffusers \cite{von-platen-etal-2022-diffusers}. We use the scheduler DPMSolver\cite{lu2022dpm}, for faster generation, to step from $\bm z_t$ to $\bm z_{t-1}$, with $50$ timesteps for generation. We use a dataset composed of images from the Landscape dataset\cite{esser2021taming} and the Kodak dataset\cite{Kodak}, images were cropped to fit in $512\times 512$, but our method can easily be adapted to any image resolution.

We compare our fine color guidance with universal guidance applied to color and initialized diffusion from \cite{meng2021sdedit}. We use a color guidance scale of $1$ for our method, using only the determined parameters. In the case of universal guidance, we compute the best coefficient $\lambda$ in their formula \eqref{eq:univ} by using our formulation of guidance, we verify experimentally its coherence. For initialized diffusion, we start the generation at 55\% of the total timesteps to maximize the information on the color map given without impairing the generative quality too much, as explained in \cite{meng2021sdedit}. We show the performance of the different methods in Fig. \ref{fig:color_control}. The images generated using fine guidance systematically follow closely the given color map. On the other hand, the two other methods drift away from the original image, losing some information given by the color map. With universal guidance being slightly better than the initialized diffusion. With a quantitative evaluation presented in Table \ref{tab:color_scores}, we confirm those results. Using the MSE between the original and generated image as criteria, we notice that our method preserves most of the information. We also confirm using image quality assessment metric (IQA) that, while initialized diffusion comes at the cost of the quality of images, using universal guidance or fine guidance does not degrade the image. The quality score of the generated images is similar to natural clean images of the dataset. For more information on the quality of generated images, FID of such models can be found in the paper of the latent diffusion models \cite{rombach_high-resolution_2022}, but were not measured here because of the size of the dataset considered.

\begin{table}[htbp]
\caption{Evaluation of color control}
\begin{center}
    \begin{tabular}{lcccc}
        \toprule
                                            & \multicolumn{1}{c}{Color fidelity} $\downarrow$ & \multicolumn{2}{c}{Image realism} $\uparrow$ \\
                                            & \textbf{MSE}  & \textbf{CLIPIQA\cite{wang2023exploring}} & \textbf{DBCNN\cite{zhang2018blind}}\\
        \midrule
        color map                           & 0.009 &  -   &   -  \\
        \midrule
        Initialized\cite{meng2021sdedit}    & 0.080 & 0.49 & 0.48 \\
        Universal\cite{bansal2023universal} & 0.031 & 0.69 & 0.66 \\
        Ours                                & 0.015 & 0.66 & 0.63 \\
        \bottomrule
    \end{tabular}
    \label{tab:color_scores}
\end{center}
\end{table}

Fine guidance can be applied to different color map resolutions. We show in Fig. \ref{fig:other_res} a comparison with universal guidance, on $16\times16$ and $8\times8$ color maps. With less information, guidance constraints are relaxed, and the difference between methods is less noticeable. We can see that the universal guidance does not use the additional information provided by a higher resolution color maps, indeed, the images generated using $8\times8$ color maps are quite similar to the ones generated using higher resolutions, this affirmation is later confirmed in the next section. We can also see, from the lower resolution generated images, that the semantic information becomes crucial when the color maps are not precise enough, such as Eiffel Tower color map in $8\times 8$. We can no longer see the semantic information from the color map only, and semantic is required to get an image close to the original.

\subsubsection{Discussion}

Initializing the diffusion is a simple method that does not require a computation of gradients, however, this is a method that loses both in realism of generation and fidelity to the color map because of the presence of a tradeoff between both. Our formulation of guidance is similar to the formulation of universal guidance, as both coincidentally requires computing similar gradients. The main difference between the two methods lies in the scaling of the loss. While in universal, the importance of the guide decreases after each step, we found that to optimally guide the scaling should not decrease in $\sqrt{1-\alpha_t}$ but remain high instead because of the ratio $\sqrt{\alpha_t}/\Bar{\lambda_t}$, the generated results confirm this observation.

\subsection{Fine guidance for compression}

We apply fine color guidance in the compression framework previously described in Section \ref{sect:compression}. The color map and the CLIP latent are quantized to reach lower bitrate. The clip latent is quantified using $1$ bit per dimension, and the color maps' palette are quantified over $5$ bits per channel as in the CoCliCo pipeline. We use the fine color guidance to preserve most of the information that is sent inside the color map. We compare our method to state-of-the-art compression methods, notably, VVC. But we also compare with other generative approach aiming at semantic compression rather than at the pixel-value fidelity, such as Text+sketch\cite{lei2023text+}. We consider the generation of images in patch of $512\times512$. 

We evaluate our method on two configurations. One using only a $16\times16$ color map resolution, targeting a bitrate similar to the lowest VVC quantization parameter. And another one, keeping more information on the colors, using a $25\times25$ color map, targeting a bitrate similar to text+sketch. For the $25\times 25$ configuration, we compare our method to Text+Sketch, their choice of representation uses similar in bitrate. For VVC, we use the intra coder(v1.6) with Vvenc implementation \cite{VVenC}, and we target similar bitrate. Images in Fig. \ref{fig:compression25} showcase the different compression framework, because the focus of VVC is an optimization of PSNR, the quality of the reconstructed image is lacking. Images reconstructed using a text and sketch semantic representation have a distinctly different aspect because of the incoherence of color, sometimes mistaking the sky for something else in the reconstruction. We show that furthermore in Fig. \ref{fig:big_compression} that images generated using our framework can have a high quality of texture on top of following the color map. This offers clear advantages over classical methods like VVC.

With the other configuration in $16\times16$, showed in Fig. \ref{fig:compression16}, at those rates, the classical approach cannot always preserve information of the color as seen in the first and fifth row. The reconstructed images are still faithful in terms of semantic, rather losing fidelity to the input in terms of color and thus PSNR. 

\begin{table*}[t]
    \centering
    \caption{Compression at extremely low bitrates, *$=$proposed approach}
    \begin{tabular}{lccccc}
        \toprule
                                                     & \multicolumn{2}{c}{$\Psi$} $\downarrow$  & \multicolumn{2}{c}{$\Phi$} $\uparrow$     &  $R$ $\downarrow$         \\ 
                                                     \cmidrule(lr){2-3}\cmidrule(lr){4-5}\cmidrule(lr){6-6}
                                                     & \textbf{MSE} & \textbf{CLIP\cite{radford2021learning}} & \textbf{CLIPIQA\cite{wang2023exploring}} & \textbf{DBCNN\cite{zhang2018blind}}  & \textbf{bits}\\
        \midrule
        \midrule
        Original images                              &   -   & 0.00 & 0.70 & 0.63 &  -    \\
        \midrule
        VVC (qp 63)                                  & 0.011 & 0.41 & 0.20 & 0.20 & 3220  \\
        CLIP+Initialized($16\times 16$)              & 0.091 & 0.21 & 0.56 & 0.50 & 2688  \\
        CLIP+Universal($16\times16$)                 & 0.034 & 0.18 & 0.64 & 0.60 & 2688  \\
        CLIP+Fine guidance*($16\times 16$)           & 0.021 & 0.18 & 0.62 & 0.58 & 2688  \\
        \midrule
        VVC (qp 55)                                  & 0.007 & 0.30 & 0.19 & 0.26 & 6640  \\
        Text+Sketch\cite{bansal2023universal}        & 0.150 & 0.19 & 0.73 & 0.61 & 6638  \\
        CLIP+Initialized($25\times25$)               & 0.035 & 0.22 & 0.55 & 0.50 & 5838  \\
        CLIP+Universal($25\times25$)                 & 0.033 & 0.20 & 0.64 & 0.60 & 5838  \\
        CLIP+Fine guidance*($25\times 25$)           & 0.016 & 0.16 & 0.64 & 0.60 & 5838  \\
        \bottomrule
    \end{tabular}
    \label{tab:comp_scores}
\end{table*}

A quantitative evaluation of the compression framework is presented in Table \ref{tab:comp_scores}. We evaluate on the two criteria presented in our formulation in \eqref{eq:form_comp} comparing on lower or equal bitrates. The first criterion, $\Psi$, is the preservation of the given semantic. We measure the preservation of the general color aspect using the MSE, and preservation of a more textual aspect of the semantic using CLIP sine similarity. Over the color preservation, we place right behind VVC on both configurations. Note that if the MSE is used as a color preservation metric here, the objective of VVC and ours do not align, indeed, we do not aim at pixel fidelity only. Over the semantic aspect, our method scores higher than the others. The second criterion is the realism, $\Phi$, used for the evaluation of the guidance, we also use IQA metrics, and while text+sketch scores higher on this criterion, this is mainly due to the use of a different model rather than a performance issue caused by our guidance. Indeed, we reach a realism score which is close to the images from the dataset. IQA metrics do not necessarily reflect perfectly the notion of realism, which is hard to grasp, however, we noted that it makes a good distinction between images with a lot of artifacts, such as VVC, and clean images. The realism of generated images can also be appreciated in Fig. \ref{fig:big_compression}. On both configurations, our method manages to preserve the semantic and color information while keeping low bitrates. Our fine guidance also allows us to use higher resolution color maps without wasting the bitrate invested in it. Indeed, as observed in Fig. \ref{fig:other_res}, while our methods shows an important difference in terms of MSE when increasing the resolution of the color map, the score barely changes for other guidance methods. We show in the figure and table that the resolution of the color maps is an interesting parameter to vary. We present in the table the rate per image rather than a bit per pixel, to be independent of the diffusion model output capacity since our description of an image does not depend on its size.

The MSE is here used as a metric of color fidelity and not as an objective to optimize at all cost. The goal of guidance is to get closer to the information brought by the color map. The use of our fine color guidance in compression brings fidelity to the color map, it can be applied on the latest diffusion model, even when trained in the latent space of a VAE.

\section{Conclusion}

We proposed a method for color guidance in diffusion models, more specifically, for latent diffusion models that does not require any training. This method outperforms other training-free methods and can be applied to any latent or pixel space diffusion model. We can then advantageously use it on the latest diffusions models to control the output. We showcased the use of our method in a context of compression at extremely low bitrates, showing that a low resolution color map can be a good descriptor to add to the semantic. We show, in the context of the CoCliCo coding pipeline, that the information given by the color map is better preserved using fine color guidance, avoiding the waste of bitrate.

\section{Acknowledgements}

This work was funded by the French National Research Agency (\emph{MADARE}, Project-ANR-21-CE48-0002).

\section*{}
\bibliographystyle{IEEEtran}
\bibliography{biblio}
\end{document}